\documentclass[10pt,twocolumn,letterpaper]{article}

\usepackage{cvpr}
\usepackage{times}
\usepackage{epsfig}
\usepackage{graphicx}
\usepackage{amsmath}
\usepackage{amssymb}

\usepackage{algorithm}
\usepackage{algorithmic}
\usepackage{multirow}
\usepackage{array, booktabs}
\usepackage{arydshln}
\usepackage{calrsfs}

\usepackage{kotex}
\usepackage{color}
\DeclareMathAlphabet{\pazocal}{OMS}{zplm}{m}{n}

\definecolor{greyblue}{rgb}{0.1,0.6,0.5}

\definecolor{purple}{rgb}{0.626,0.125,0.941}
\newcommand\jh[1]{\textcolor{purple}{#1}}

\usepackage[pagebackref=true,breaklinks=true,letterpaper=true,colorlinks,bookmarks=false]{hyperref}

\cvprfinalcopy 

\def\cvprPaperID{4706} 

\ifcvprfinal\fi
\begin{document}


\title{Character Region Awareness for Text Detection}

\author{
 Youngmin Baek, Bado Lee, Dongyoon Han, Sangdoo Yun, and Hwalsuk Lee\thanks{Corresponding author.}\\
 Clova AI Research, NAVER Corp.\\
 {\tt\small\{youngmin.baek, bado.lee, dongyoon.han, sangdoo.yun, hwalsuk.lee\}@navercorp.com} \\
}

\maketitle

\begin{abstract}
Scene text detection methods based on neural networks have emerged recently and have shown promising results. Previous methods trained with rigid word-level bounding boxes exhibit limitations in representing the text region in an arbitrary shape. In this paper, we propose a new scene text detection method to effectively detect text area by exploring each character and affinity between characters. To overcome the lack of individual character level annotations, our proposed framework exploits both the given character-level annotations for synthetic images and the estimated character-level ground-truths for real images acquired by the learned interim model. In order to estimate affinity between characters, the network is trained with the newly proposed representation for affinity. Extensive experiments on six benchmarks, including the TotalText and CTW-1500 datasets which contain highly curved texts in natural images, demonstrate that our character-level text detection significantly outperforms the state-of-the-art detectors. According to the results, our proposed method guarantees high flexibility in detecting complicated scene text images, such as arbitrarily-oriented, curved, or deformed texts.
\end{abstract}

\section{Introduction}

Scene text detection has attracted much attention in the computer vision field because of its numerous applications, such as instant translation, image retrieval, scene parsing, geo-location, and blind-navigation. Recently, scene text detectors based on deep learning have shown promising performance~\cite{he2017single,zhou2017east,liu2018fots,deng2018pixellink,he2017deep,he2018end,hu2017wordsup,jiang2017r2cnn,liao2018textboxes++,long2018textsnake,lyu2018mask,shi2017detecting,lyu2018multi}.
These methods mainly train their networks to localize word-level bounding boxes. However, they may suffer in difficult cases, such as texts that are curved, deformed, or extremely long, which are hard to detect with a single bounding box. Alternatively, character-level awareness has many advantages when handling challenging texts by linking the successive characters in a bottom-up manner. Unfortunately, most of the existing text datasets do not provide character-level annotations, and the work needed to obtain character-level ground truths is too costly. 

\begin{figure}[t]
  \centering
  \includegraphics*[width=0.9\linewidth, clip=true]{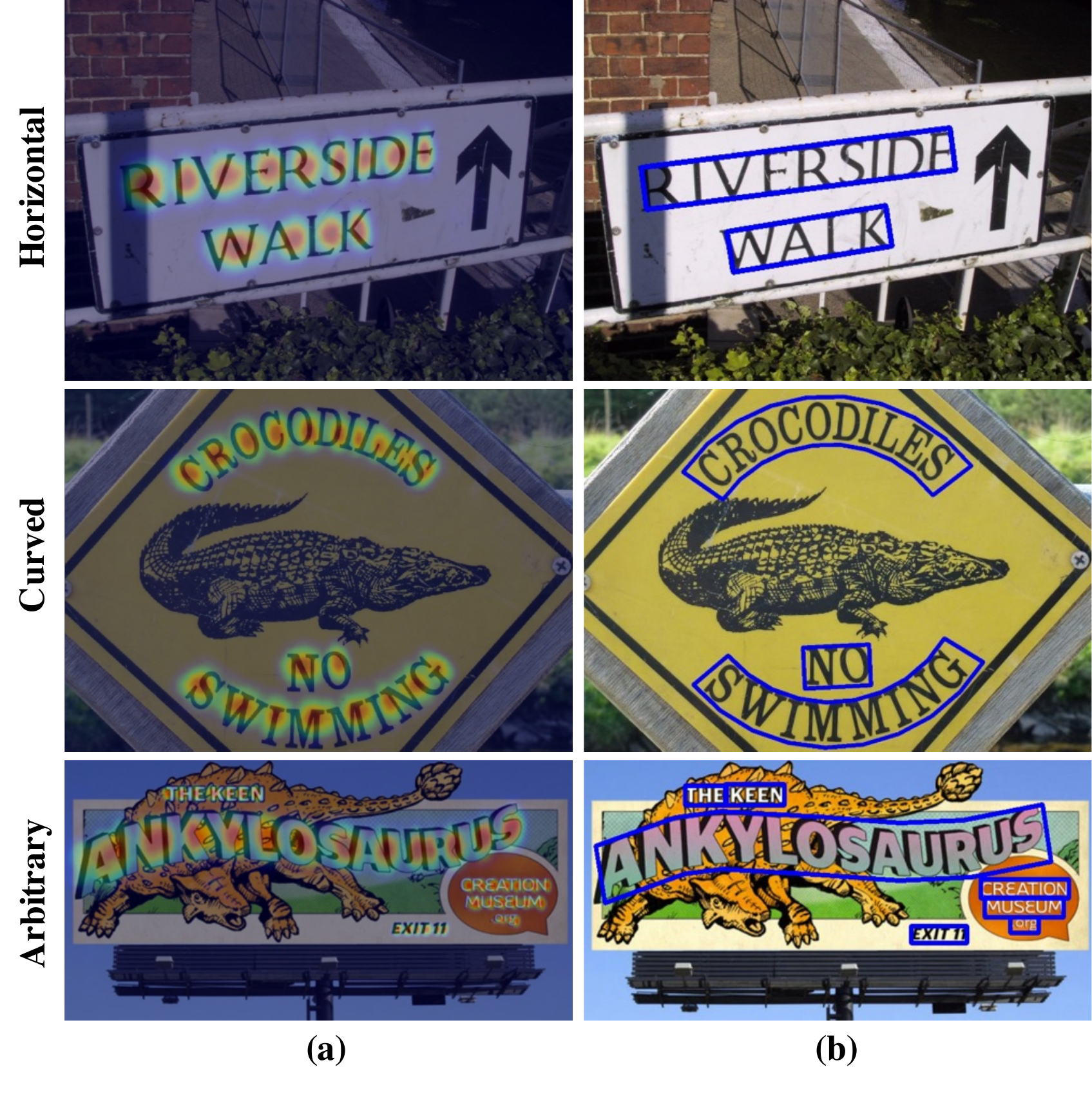}
  \caption{Visualization of character-level detection using CRAFT. (a) Heatmaps predicted by our proposed framework. (b) Detection results for texts of various shape.}
  \label{fig:overview} 
  \vspace{-3mm}
\end{figure}

In this paper, we propose a novel text detector that localizes the individual character regions and links the detected characters to a text instance. Our framework, referred to as CRAFT for \textit{Character Region Awareness For Text detection}, is designed with a convolutional neural network producing the character \textit{region score} and \textit{affinity score}. The \textit{region score} is used to localize individual characters in the image, and the \textit{affinity score} is used to group each character into a single instance. To compensate for the lack of character-level annotations, we propose a weakly-supervised learning framework that estimates character-level ground truths in existing real word-level datasets.

Figure.~\ref{fig:overview} is a visualization of CRAFT's results on various shaped texts. By exploiting character-level region awareness, texts in various shapes are easily represented. We demonstrate extensive experiments on ICDAR datasets~\cite{karatzas2013icdar,karatzas2015icdar,nayef2017icdar2017} to validate our method, and the experiments show that the proposed method outperforms state-of-the-art text detectors. Furthermore, experiments on MSRA-TD500, CTW-1500, and TotalText datasets~\cite{yao2012detecting,yuliang2017detecting,ch2017total} show the high flexibility of the proposed method on complicated cases\jh{,} such as long, curved, and/or arbitrarily shaped texts.


\section{Related Work}

The major trend in scene text detection before the emergence of deep learning was bottom-up, where handcrafted features were mostly used -- such as MSER~\cite{matas2004robust} or SWT~\cite{epshtein2010detecting}-- as a basic component. 
Recently, deep learning-based text detectors have been proposed by adopting popular object detection/segmentation methods like SSD~\cite{liu2016ssd}, Faster R-CNN~\cite{ren2017faster}, and FCN~\cite{long2015fully}. 

\vspace{2mm}
{\bf Regression-based text detectors }
Various text detectors using box regression adapted from popular object detectors have been proposed. Unlike objects in general, texts are often presented in irregular shapes with various aspect ratios. To handle this problem, TextBoxes~\cite{liao2017textboxes} modified convolutional kernels and anchor boxes to effectively capture various text shapes. DMPNet~\cite{liu2017deep} tried to further reduce the problem by incorporating quadrilateral sliding windows. In recent, Rotation-Sensitive Regression Detector (RSDD) \cite{liao2018rotation} which makes full use of rotation-invariant features by actively rotating the convolutional filters was proposed. However, there is a structural limitation to capturing all possible shapes that exist in the wild when using this approach.

\vspace{2mm}
{\bf Segmentation-based text detectors } 
Another common approach is based on works dealing with segmentation, which aims to seek text regions at the pixel level. These approaches that detect texts by estimating word bounding areas, such as Multi-scale FCN~\cite{he2017multi}, Holistic-prediction~\cite{yao2016scene}, and PixelLink~\cite{deng2018pixellink} have also been proposed using segmentation as their basis. 
SSTD~\cite{he2017single} tried to benefit from both the regression and segmentation approaches by using an attention mechanism to enhance text related area via reducing background interference on the feature level. Recently, TextSnake~\cite{long2018textsnake} was proposed to detect text instances by predicting the text region and the center line together with geometry attributes.



\begin{figure}[t!]
	\begin{center}
        \includegraphics*[width=0.9\linewidth, clip=true]{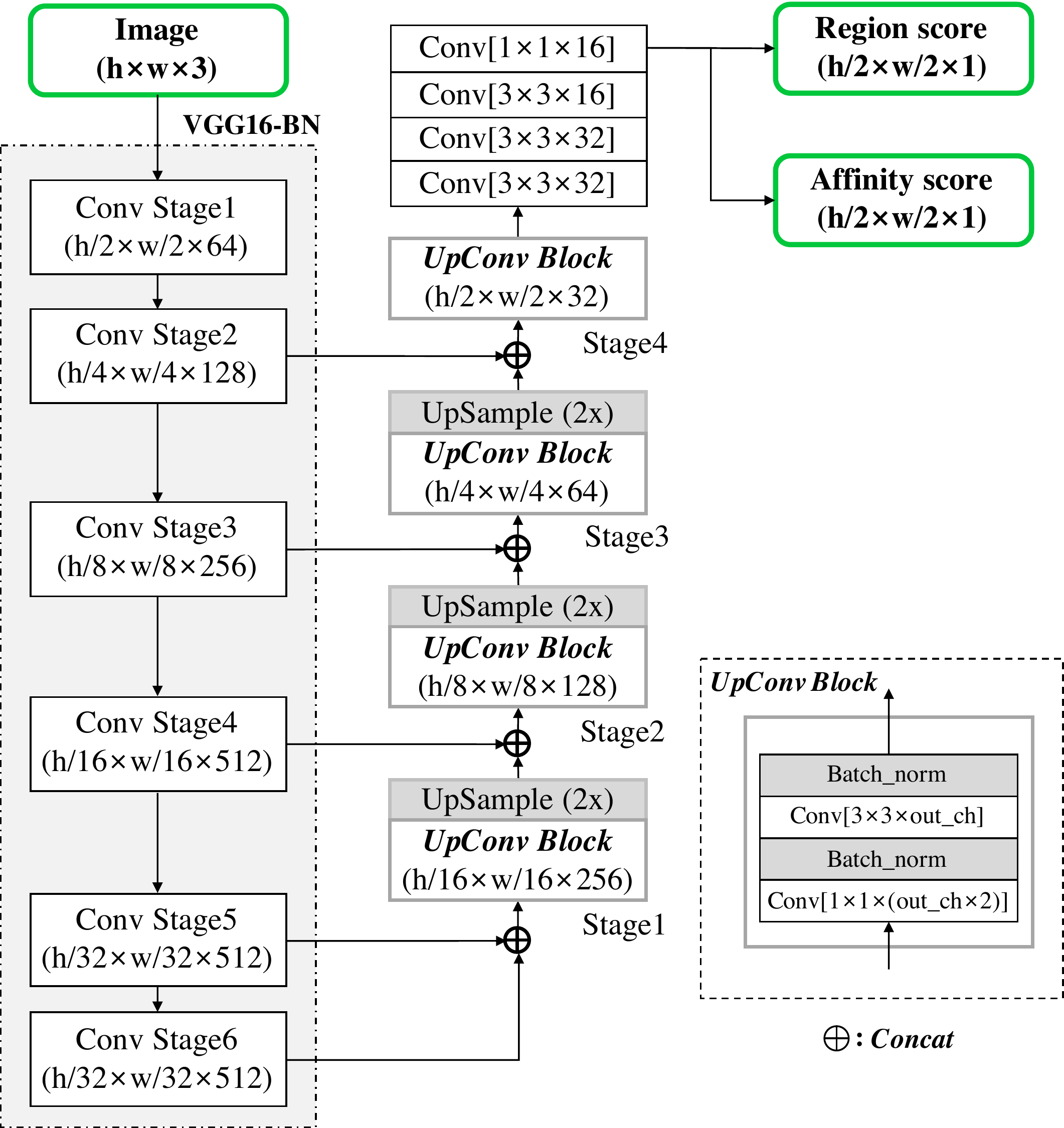}
         \vspace{+3mm}
        \caption{Schematic illustration of our network architecture.}
  	    \label{fig:architecture} 
    \end{center}
     \vspace{-7.5mm}
\end{figure}

\begin{figure*}[t!]
  \begin{center}
    \includegraphics*[width=0.9\linewidth, clip=true]{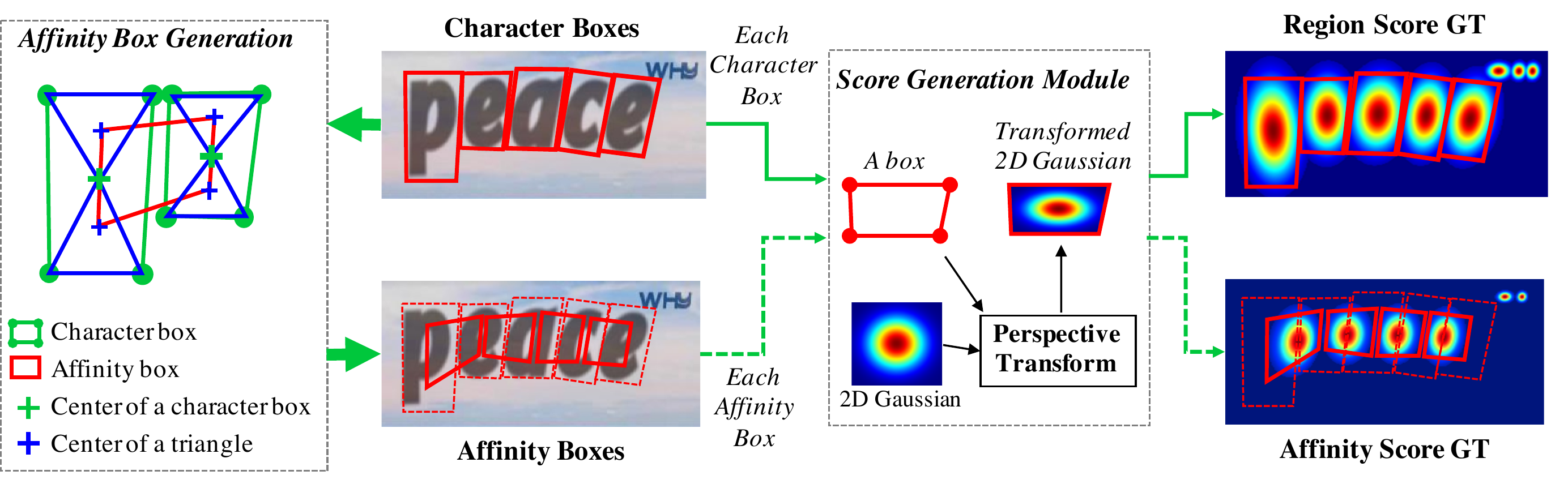}
    \vspace{+2mm}
    \caption{Illustration of ground truth generation procedure in our framework. We generate ground truth labels from a synthetic image that has character level annotations.}
  \label{fig:gtgeneration} 
  \end{center}
  \vspace{-3mm}
\end{figure*}

\vspace{2mm}
{\bf End-to-end text detectors } An end-to-end approach trains the detection and recognition modules simultaneously so as to enhance detection accuracy by leveraging the recognition result. FOTS~\cite{liu2018fots} and EAA~\cite{he2018end} concatenate popular detection and recognition methods, and train them in an end-to-end manner.
Mask TextSpotter~\cite{lyu2018mask} took advantage of their unified model to treat the recognition task as a semantic segmentation problem. 
It is obvious that training with the recognition module helps the text detector be more robust to text-like background clutters.

Most methods detect text with words as its unit, but defining the extents to a word for detection is non-trivial since words can be separated by various criteria, such as meaning, spaces or color. In addition, the boundary of the word segmentation cannot be strictly defined, so the word segment itself has no distinct semantic meaning. This ambiguity in the word annotation dilutes the meaning of the ground truth for both regression and segmentation approaches. 


{\bf Character-level text detectors } Zhang et al.~\cite{zhang2016multi} proposed a character level detector using text block candidates distilled by MSER~\cite{matas2004robust}. The fact that it uses MSER to identify individual characters limits its detection robustness under certain situations, such as scenes with low contrast, curvature, and light reflection. 
Yao et al.~\cite{yao2016scene} used a prediction map of the characters along with a map of text word regions and linking orientations that require character level annotations. 
Instead of an explicit character level prediction, Seglink ~\cite{shi2017detecting} hunts for text grids (partial text segments) and associates these segments with an additional link prediction.
Even though Mask TextSpotter~\cite{lyu2018mask} predicts a character-level probability map, it was used for text recognition instead of spotting individual characters.

This work is inspired by the idea of WordSup~\cite{hu2017wordsup}, which uses a weakly supervised framework to train the character-level detector. However, a disadvantage of Wordsup is that the character representation is formed in rectangular anchors, making it vulnerable to perspective deformation of characters induced by varying camera viewpoints. Moreover, it is bound by the performance of the backbone structure (i.e. using SSD and being limited by the number of anchor boxes and their sizes).

%



\section{Methodology}

Our main objective is to precisely localize each individual character in natural images. To this end, we train a deep neural network to predict character regions and the affinity between characters. Since there is no public character-level dataset available, the model is trained in a weakly-supervised manner.

\subsection{Architecture}
A fully convolutional network architecture based on VGG-16~\cite{simonyan2014very} with batch normalization is adopted as our backbone. Our model has skip connections in the decoding part, which is similar to U-net~\cite{ronneberger2015u} in that it aggregates low-level features. The final output has two channels as score maps: the \textit{region score} and the \textit{affinity score}.
The network architecture is schematically illustrated in Fig.~\ref{fig:architecture}.


\subsection{Training} 

\subsubsection{Ground Truth Label Generation}

For each training image, we generate the ground truth label for the \textit{region score} and the \textit{affinity score} with character-level bounding boxes. 
The \textit{region score} represents the probability that the given pixel is the center of the character, and the \textit{affinity score} represents the center probability of the space between adjacent characters. 

Unlike a binary segmentation map, which labels each pixel discretely, we encode the probability of the character center with a Gaussian heatmap.
This heatmap representation has been used in other applications, such as in pose estimation works~\cite{cao2017realtime,newell2016stacked} due to its high flexibility when dealing with ground truth regions that are not rigidly bounded. We use the heatmap representation to learn both the \textit{region score} and the \textit{affinity score}.

Fig.~\ref{fig:gtgeneration} summarizes the label generation pipeline for a synthetic image. Computing the Gaussian distribution value directly for each pixel within the bounding box is very time-consuming. Since character bounding boxes on an image are generally distorted via perspective projections, we use the following steps to approximate and generate the ground truth for both the \textit{region score} and the \textit{affinity score}: 
1) prepare a 2-dimensional isotropic Gaussian map;
2) compute perspective transform between the Gaussian map region and each character box;
3) warp Gaussian map to the box area.

For the ground truths of the \textit{affinity score}, the affinity boxes are defined using adjacent character boxes, as shown in Fig.~\ref{fig:gtgeneration}.
By drawing diagonal lines to connect opposite corners of each character box, we can generate two triangles -- which we will refer to as the upper and lower character triangles.
Then, for each adjacent character box pair, an affinity box is generated by setting the centers of the upper and lower triangles as corners of the box.


The proposed ground truth definition enables the model to detect large or long-length text instances sufficiently, despite using small receptive fields. On the other hand, previous approaches like box regression require a large receptive field in such cases. Our character-level detection makes it possible for convolutional filters to focus only on intra-character and inter-character, instead of the entire text instance.

\begin{figure*}[t]
  \centering
  \includegraphics[width=0.98\linewidth]{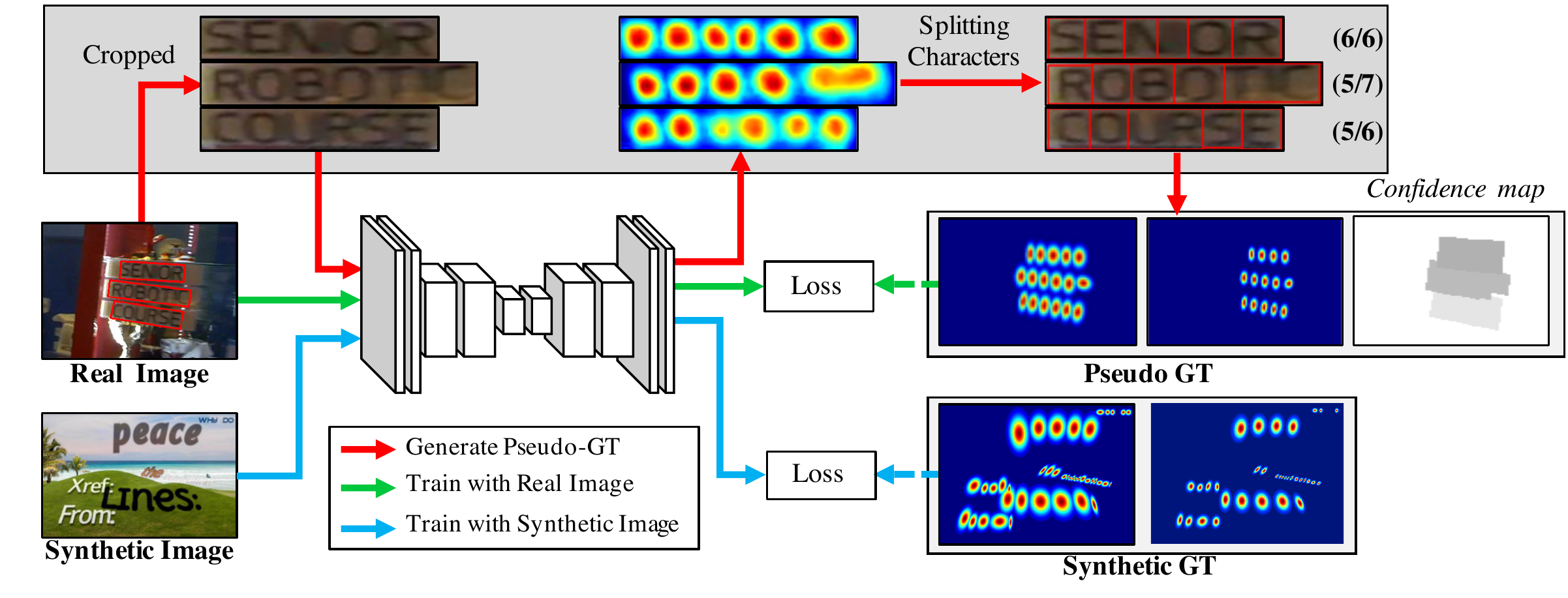}
  \caption{Illustration of the overall training stream for the proposed method. Training is carried out using both real and synthetic images in a weakly-supervised fashion.}
\label{fig:weaklysupervsion} 
\end{figure*}

\subsubsection{Weakly-Supervised Learning}
Unlike synthetic datasets, real images in a dataset usually have word-level annotations. Here, we generate character boxes from each word-level annotation in a weakly-supervised manner, as summarized in Fig.~\ref{fig:weaklysupervsion}. When a real image with word-level annotations is provided, the learned interim model predicts the character region score of the cropped word images to generate character-level bounding boxes. In order to reflect the reliability of the interim model's prediction, the value of the confidence map over each word box is computed proportional to the number of the detected characters divided by the number of the ground truth characters, which is used for the learning weight during training.

\begin{figure*}[h!]
  \centering
  \includegraphics*[width=0.95\linewidth, clip=true]{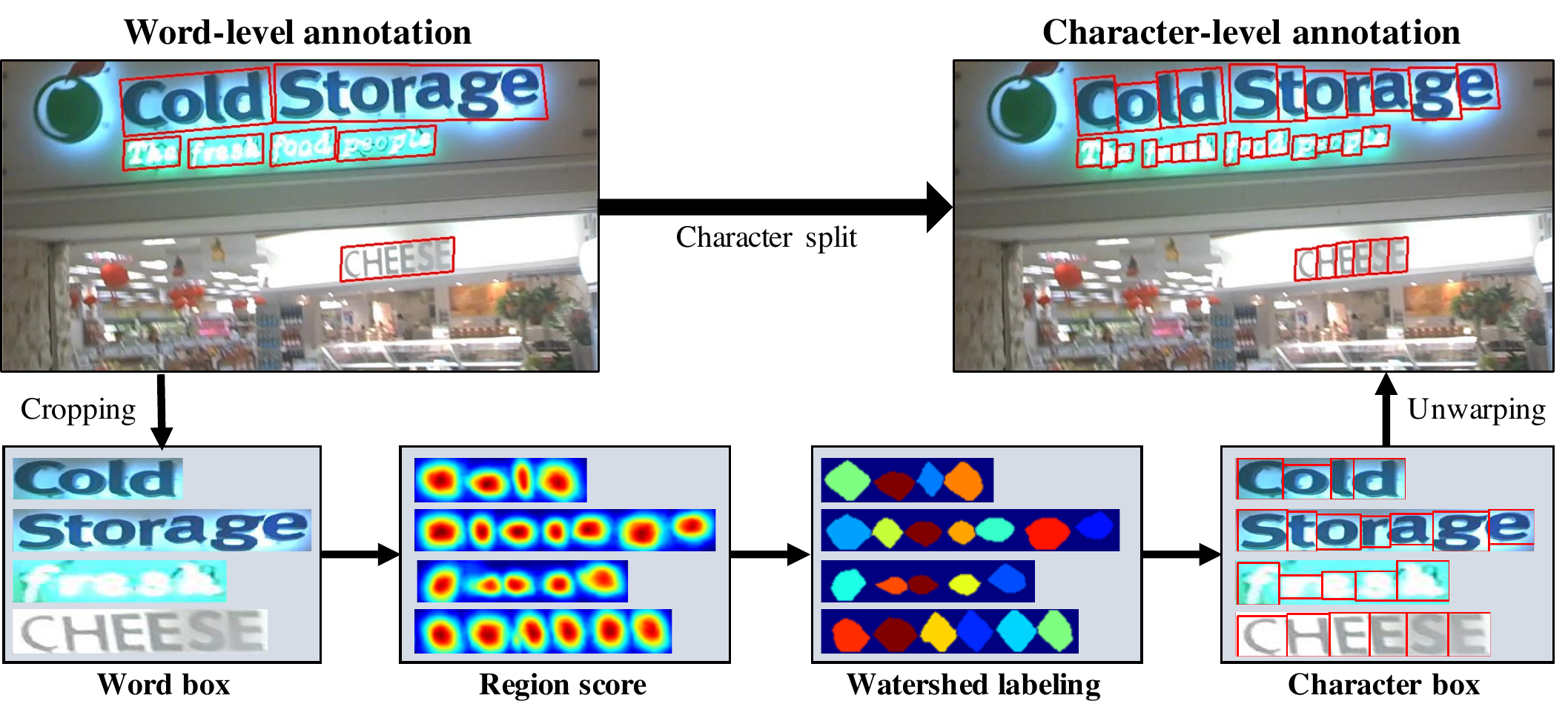}
  \caption{Character split procedure for achieving character-level annotation from word-level annotation: 1) crop the word-level image; 2) predict the region score; 3) apply the watershed algorithm; 4) get the character bounding boxes; 5) unwarp the character bounding boxes.} 
  \label{fig:charactersplit} 
\end{figure*}

Fig.~\ref{fig:charactersplit} shows the entire procedure for splitting the characters. First, the word-level images are cropped from the original image. Second, the model trained up to date predicts the \textit{region score}. Third, the watershed algorithm~\cite{vincent1991watersheds} is used to split the character regions, which is used to make the character bounding boxes covering regions. Finally, the coordinates of the character boxes are transformed back into the original image coordinates using the inverse transform from the cropping step. The pseudo-ground truths (pseudo-GTs) for the \textit{region score} and the \textit{affinity score} can be generated by the steps described in Fig.~\ref{fig:gtgeneration} using the obtained quadrilateral character-level bounding boxes.


When the model is trained using weak-supervision, we are compelled to train with incomplete pseudo-GTs. 
If the model is trained with inaccurate region scores, the output might be blurred within character regions. 
To prevent this, we measure the quality of each pseudo-GTs generated by the model.
Fortunately, there is a very strong cue in the text annotation, which is the \textit{word length}. 
In most datasets, the transcription of words is provided and the length of the words can be used to evaluate the confidence of the pseudo-GTs.

\begin{figure}[t]
  \includegraphics*[width=8.3cm, clip=true]{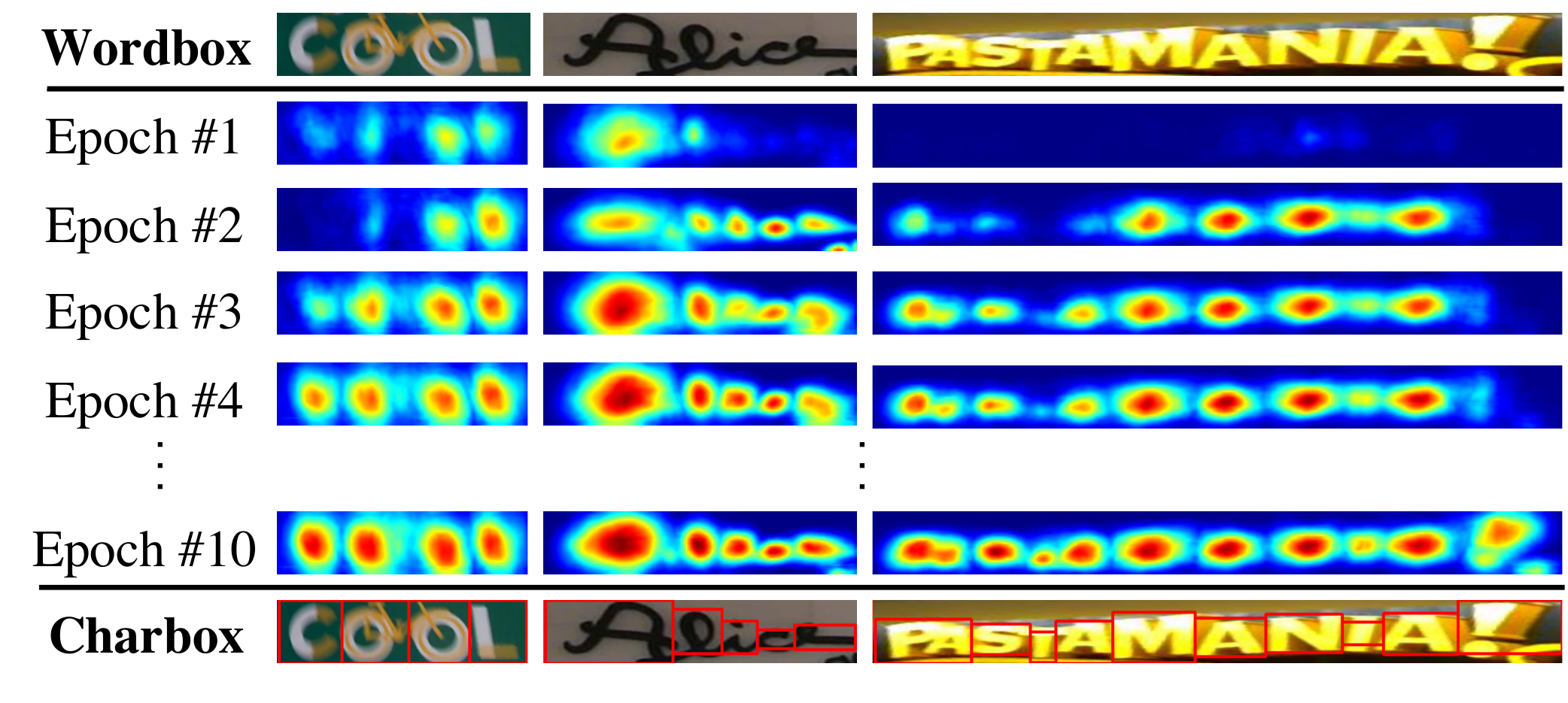}
  \caption{Character region score maps during training.}
  \label{fig:scoremapduringtraining} 
\end{figure}

For a word-level annotated sample $w$ of the training data, let $R(w)$ and $l(w)$ be the bounding box region and the word length of the sample $w$, respectively.
Through the character splitting process, we can obtain the estimated character bounding boxes and their corresponding length of characters $l^c(w)$.
Then the confidence score $s_{conf}(w)$ for the sample $w$ is computed as,

\begin{equation} \label{eq:lr}
    s_{conf}(w) = \frac{l(w) - \min(l(w), |l(w)-l^c(w)|)}{l(w)},
\end{equation}
and the pixel-wise confidence map $S_c$ for an image is computed as, 
\begin{equation}
    S_c(p) =
    \begin{cases}
      s_{conf}(w) & p \in R(w), \\
      1 & \text{otherwise},
    \end{cases}
\end{equation}
where $p$ denotes the pixel in the region $R(w)$. 
The objective $L$ is defined as,
\begin{equation} \label{eq:weighted_loss}
\begin{aligned}
L & = \hspace{-1mm} \sum_{p}S_c(p) \hspace{-0.5mm} \cdot \hspace{-0.5mm} \left( ||S_{r}(p)-S^{*}_{r}(p)||^{2}_{2} \hspace{-0.3mm} + \hspace{-0.3mm} ||S_{a}(p)-S^{*}_{a}(p)||^{2}_{2} \right),
\end{aligned}
\end{equation}
where $S_r^{*}(p)$ and $S_a^{*}(p)$ denote the pseudo-ground truth \textit{region score} and \textit{affinity map}, respectively, and $S_r(p)$ and $S_a(p)$ denote the predicted \textit{region score} and \textit{affinity score}, respectively.
When training with synthetic data, we can obtain the real ground truth, so $S_c(p)$ is set to $1$.

As training is performed, the CRAFT model can predict characters more accurately, and the confidence scores $s_{conf}(w)$ are gradually increased as well. 
Fig.~\ref{fig:scoremapduringtraining} shows the character region score map during training. At the early stages of training, the region scores are relatively low for unfamiliar text in natural images. The model learns the appearances of new texts, such as irregular fonts, and synthesized texts that have a different data distribution against that of the SynthText dataset. 

If the confidence score $s_{conf}(w)$ is below $0.5$, the estimated character bounding boxes should be neglected since they have adverse effects when training the model. In this case, we assume the width of the individual character is constant and compute the character-level predictions by simply dividing the word region $R(w)$ by the number of characters $l(w)$. Then, $s_{conf}(w)$ is set to $0.5$ to learn unseen appearances of texts.



\subsection{Inference}
At the inference stage, the final output can be delivered in various shapes, such as word boxes or character boxes, and further polygons.
For datasets like ICDAR, the evaluation protocol is word-level intersection-over-union (IoU), so here we describe how to make word-level bounding boxes \textit{QuadBox} from the predicted $S_r$ and $S_a$ through a simple yet effective post-processing step. 

The post-processing for finding bounding boxes is summarized as follows. First, the binary map $M$ covering the image is initialized with 0. ${M}(p)$ is set to 1 if $S_{r}(p) > {\tau}_{r}$ or $S_{a}(p) > {\tau}_{a}$, where ${\tau}_{r}$ is the region threshold and ${\tau}_{a}$ is the affinity threshold. Second, Connected Component Labeling (CCL) on ${M}$ is performed. Lastly, \textit{QuadBox} is obtained by finding a rotated rectangle with the minimum area enclosing the connected components corresponding to each of the labels. The functions like \textit{connectedComponents} and \textit{minAreaRect} provided by OpenCV can be applied for this purpose.



Note that an advantage of CRAFT is that it does not need any further post-processing methods, like Non-Maximum Suppression (NMS). Since we have image blobs of word regions separated by CCL, the bounding box for a word is simply defined by the single enclosing rectangle. On a different note, our character linking process is conducted at a pixel-level, which differs from other linking-based methods~\cite{shi2017detecting,hu2017wordsup} relying on searching relations between text components explicitly.

\begin{figure}[t]
 \begin{center}
   \includegraphics*[width=7.8cm, clip=true]{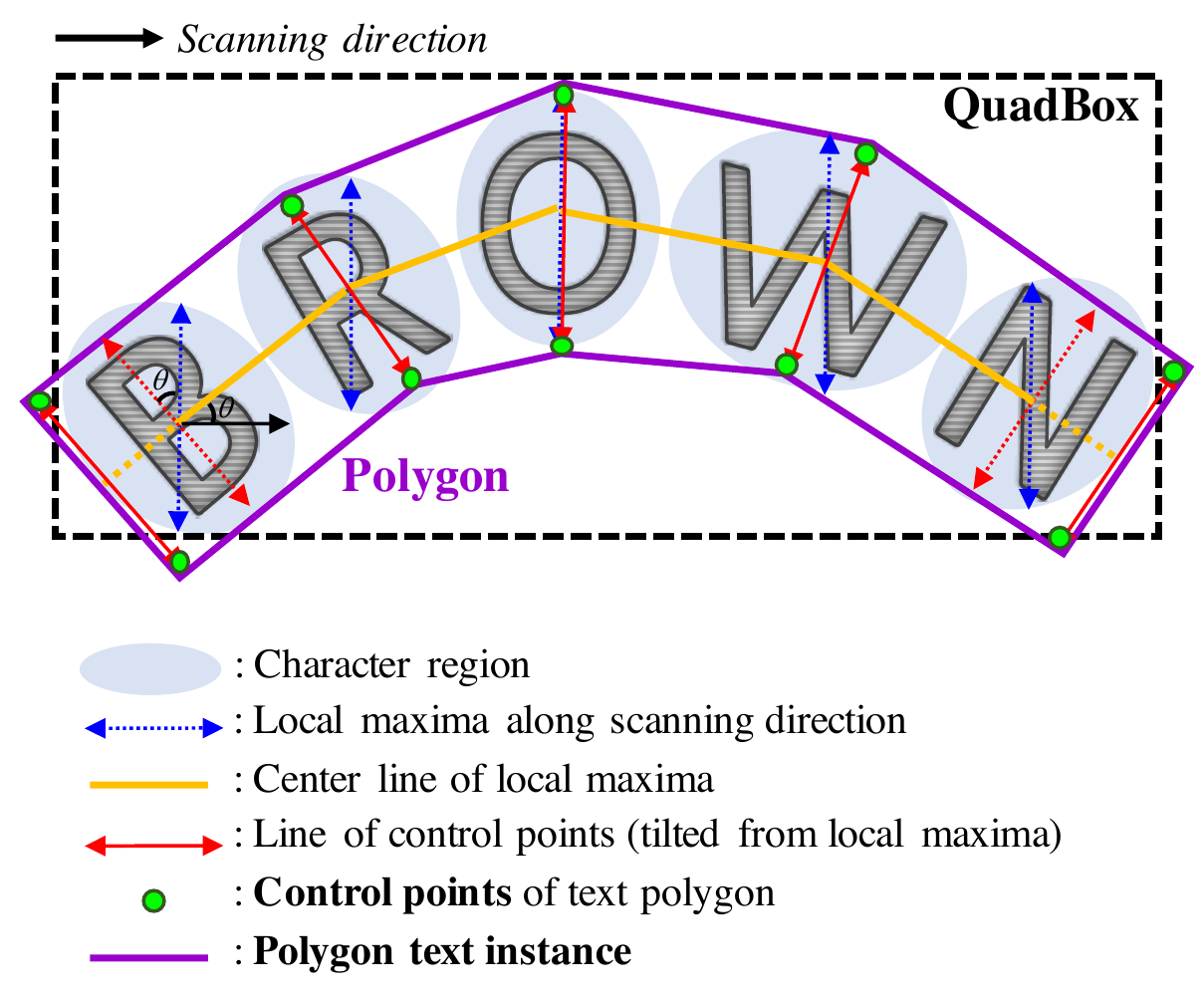}
   \vspace{-1mm}
   \caption{Polygon generation for arbitrarily-shaped texts.}
   \label{fig:poly_postprocessing} 
   \vspace{-5mm}
  \end{center}
\end{figure}

\begin{table*}[t!]
  \centering
  \tabcolsep=0.2cm
  \fontsize{10}{10}\selectfont
  \renewcommand*{\arraystretch}{1.1}
  \begin{tabular}{c||c|c|c||c|c|c||c|c|c||c|c|c||c}
    \hline 
    \rule{0pt}{10pt} \multirow{2}{*}{\textbf{Method}} & \multicolumn{3}{c||}{\textbf{IC13}(DetEval)} & \multicolumn{3}{c||}{\textbf{IC15}} & \multicolumn{3}{c||}{\textbf{IC17}} & \multicolumn{3}{c||}{\textbf{MSRA-TD500}} & \multirow{2}{*}{\textbf{FPS}}\\
    \cline{2-13}
    \rule{0pt}{10pt} & \textbf{R} & \textbf{P} & \textbf{H} & \textbf{R} & \textbf{P} & \textbf{H} & \textbf{R} & \textbf{P} & \textbf{H} & \textbf{R} & \textbf{P} & \textbf{H} &\\
    \hline
    \hline

    \rule{0pt}{10pt}
    Zhang et al.~\cite{zhang2016multi} & 78 & 88 & 83 & 43 & 71 & 54 & - & - & - & 67 & 83 & 74 & 0.48 \\ 
    Yao et al.~\cite{yao2016scene} & 80.2 & 88.8 & 84.3 & 58.7 & 72.3 & 64.8 & - & - & - & 75.3 & 76.5 & 75.9 & 1.61 \\ 
    SegLink~\cite{shi2017detecting}&83.0&87.7&85.3&76.8&73.1&75.0&-&-&- & 70 & 86 & 77 & 20.6\\
    SSTD~\cite{he2017single}&86&89&88&73&80&77&-&-&- & - & - & - & 7.7 \\
    Wordsup~\cite{hu2017wordsup}&87.5&93.3&90.3&77.0&79.3&78.2&-&-&- & - & - & - & 1.9\\
    EAST$^*$~\cite{zhou2017east}&-&-&-&78.3&83.3&80.7&-&-&- & 67.4 & 87.3 & 76.1 & 13.2\\
    He et al.~\cite{he2017deep} & 81 & 92 & 86 & 80 & 82 & 81 & - & - & - & 70 & 77 & 74 & 1.1 \\
    R2CNN~\cite{jiang2017r2cnn} &82.6&93.6&87.7&79.7&85.6&82.5&-&-&- & - & - & - & 0.4\\ 
    TextSnake~\cite{long2018textsnake} & - & - & - & 80.4 & 84.9 & 82.6 & - & - & - & 73.9 & 83.2 &	78.3 & 1.1 \\
    TextBoxes++$^*$~\cite{liao2018textboxes++} & 86 & 92 & 89 & 78.5 & 87.8 & 82.9 & - & - & - & - & - & - & 2.3\\ 
    \textit{EAA}~\cite{he2018end} & \textit{87} & \textit{88} & \textit{88} & \textit{83} & \textit{84} & \textit{83} & - & - & - & - & - & - & -\\ 
    \textit{Mask TextSpotter}~\cite{lyu2018mask} & \textit{88.1} & \textit{94.1} & \textit{91.0} & \textit{81.2} & \textit{85.8} & \textit{83.4} & - & -&  - & - & - & - & 4.8\\ 
    PixelLink$^*$~\cite{deng2018pixellink} & 87.5 & 88.6 & 88.1 & 82.0 & 85.5 & 83.7 & - & - & - & 73.2 &  83.0 &	77.8 &	3.0 \\
    RRD$^*$~\cite{liao2018rotation} & 86 & 92 & 89 & 80.0 & 88.0 & 83.8 & - & - & - & 73 & 87 & 79 & 10\\ 
    Lyu et al.$^*$~\cite{lyu2018multi} & 84.4 & 92.0 & 88.0 & 79.7 & 89.5 & 84.3 & \textcolor{black}{\textbf{70.6}} & 74.3 & 72.4 & 76.2 & 87.6 & 81.5 & 5.7\\ 
    \textit{FOTS}~\cite{liu2018fots} & - & - & \textit{87.3} & \textit{82.0} & \textit{88.8} & \textit{85.3} & \textit{57.5} & \textit{79.5} & \textit{66.7} & - & - & - & 23.9\\ 
    \hline
    \hline
    \rule{0pt}{10pt} \textbf{CRAFT(ours)} & \textcolor{black}{\textbf{93.1}} & \textcolor{black}{\textbf{97.4}} & \textcolor{black}{\textbf{95.2}} & \textcolor{black}{\textbf{84.3}} & \textcolor{black}{\textbf{89.8}} & \textcolor{black}{\textbf{86.9}} & 68.2 & \textcolor{black}{\textbf{80.6}} & \textcolor{black}{\textbf{73.9}} & \textcolor{black}{\textbf{78.2}} & \textcolor{black}{\textbf{88.2}} & \textcolor{black}{\textbf{82.9}} & 8.6\\
    \hline
  \end{tabular}
  \vspace{3mm}
  \caption{Results on quadrilateral-type datasets, such as ICDAR and MSRA-TD500. $^*$ denote the results based on multi-scale tests. Methods in \textit{italic} are results solely from the detection of end-to-end models for a fair comparison. R, P, and H refer to recall, precision and H-mean, respectively. The best score is highlighted in \textbf{bold}. FPS is for reference only because the experimental environments are different. 
  We report the best FPSs, each of which was reported in the original paper.}
  \label{tab:result_icdar}
\end{table*}

Additionally, we can generate a polygon around the entire character region to deal with curved texts effectively. The procedure of polygon generation is illustrated in Fig.~\ref{fig:poly_postprocessing}. 
The first step is to find the local maxima line of character regions along the scanning direction, as shown in the figure with arrows in blue. The lengths of the local maxima lines are equally set as the maximum length among them to prevent the final polygon result from becoming uneven. The line connecting all the center points of the local maxima is called the center line, shown in yellow. Then, the local maxima lines are rotated to be perpendicular to the center line to reflect the tilt angle of characters, as expressed by the red arrows. The endpoints of the local maxima lines are the candidates for the control points of the text polygon. To fully cover the text region, we move the two outer-most tilted local maxima lines outward along the local maxima center line, making the final control points (green dots).


\section{Experiment}

\subsection{Datasets}



\noindent\textbf{ICDAR2013} (IC13) was released during the ICDAR 2013 Robust Reading Competition for focused scene text detection, consisting of high-resolution images, 229 for training and 233 for testing, containing texts in English. The annotations are at word-level using rectangular boxes.


\noindent\textbf{ICDAR2015} (IC15) was introduced in the ICDAR 2015 Robust Reading Competition for incidental scene text detection, consisting of 1000 training images and 500 testing images, both with texts in English. 
The annotations are at the word level using quadrilateral boxes.


\noindent\textbf{ICDAR2017} (IC17) contains 7,200 training images, 1,800 validation images, and 9,000 testing images with texts in 9 languages for multi-lingual scene text detection. Similar to IC15, the text regions in IC17 are also annotated by the 4 vertices of quadrilaterals.


\noindent\textbf{MSRA-TD500} (TD500) contains 500 natural images, which are split into 300 training images and 200 testing images, collected both indoors and outdoors using a pocket camera. The images contain English and Chinese scripts. Text regions are annotated by rotated rectangles.

\noindent\textbf{TotalText} (TotalText), recently presented in ICDAR 2017, contains 1255 training and 300 testing images. It especially provides curved texts, which are annotated by polygons and word-level transcriptions.

\noindent\textbf{CTW-1500} (CTW) consists of 1000 training and 500 testing images. Every image has curved text instances, which are annotated by polygons with 14 vertices.


\begin{table}[t!]
  \centering
  \tabcolsep=0.12cm
  \fontsize{10}{10}\selectfont
  \renewcommand*{\arraystretch}{1.1}
  \begin{tabular}{c||c|c|c||c|c|c}
    \hline  
    \rule{0pt}{10pt} \multirow{2}{*}{\textbf{Method}} & \multicolumn{3}{c||}{\textbf{TotalText}} & \multicolumn{3}{c}{\textbf{CTW-1500}}\\
    \cline{2-7}
    \rule{0pt}{10pt} & \textbf{R} & \textbf{P} & \textbf{H} & \textbf{R} & \textbf{P} & \textbf{H}\\
    \hline
    \hline
    CTD+TLOC~\cite{yuliang2017detecting} & - & - & - & 69.8	& 77.4 & 73.4 \\
    MaskSpotter~\cite{lyu2018mask} & 55.0 & 69.0 & 61.3 & - & - & - \\
    TextSnake~\cite{long2018textsnake} &	74.5 & 82.7 & 78.4 & \textcolor{black}{\textbf{85.3}} & 67.9 & 75.6 \\
    \hline
    \hline
    \rule{0pt}{10pt} \textbf{CRAFT(ours)} & \textcolor{black}{\textbf{79.9}} & \textcolor{black}{\textbf{87.6}} & \textcolor{black}{\textbf{83.6}} & 81.1 & \textcolor{black}{\textbf{86.0}} & \textcolor{black}{\textbf{83.5}}\\
    \hline
  \end{tabular}
  \vspace{3mm}
  \caption{Results on polygon-type datasets\jh{,} such as TotalText and CTW-1500. R, P and H refer to recall, precision and H-mean\jh{,} respectively. The best score is highlighted in {\textbf{bold}}.}
  \vspace{-3mm}
  \label{tab:result_poly}
\end{table}

\subsection{Training strategy}
The training procedure includes two steps: we first use the SynthText dataset~\cite{gupta2016synthetic} to train the network for 50k iterations, then each benchmark dataset is adopted to fine-tune the model. Some “DO NOT CARE” text regions in ICDAR 2015 and ICDAR 2017 datasets are ignored in training by setting $s_{conf}(w)$ to 0. We use the ADAM~\cite{kingma2015adam} optimizer in all training processes. For multi-GPU training, the training and supervision GPUs are separated, and pseudo-GTs generated by the supervision GPUs are stored in the memory. During fine-tuning, the SynthText dataset is also used at a rate of 1:5 to make sure that the character regions are surely separated.
In order to filter out texture-like texts in natural scenes, \textit{On-line Hard Negative Mining}~\cite{shrivastava2016training} is applied at a ratio of 1:3. Also, basic data augmentation techniques like crops, rotations, and/or color variations are applied. 


Weakly-supervised training requires two types of data; quadrilateral annotations for cropping word images and transcriptions for calculating word length. The datasets meeting these conditions are IC13, IC15, and IC17. Other datasets such as MSRA-TD500, TotalText, and CTW-1500 
do not meet the requirements.
MSRA-TD500 does not provide transcriptions, while TotalText and CTW-1500 provide polygon annotations only. 
Therefore, we trained CRAFT only on the ICDAR datasets, and tested on the others without fine-tuning. Two different models are trained with the ICDAR datasets. The first model is trained on IC15 to evaluate IC15 only. The second model is trained on both IC13 and IC17 together, which is used for evaluating the other five datasets. No extra images are used for training. The number of iterations for fine-tuning is set to 25k.

\begin{figure*}[t]
  \centering
  \includegraphics*[width=17cm, clip=true]{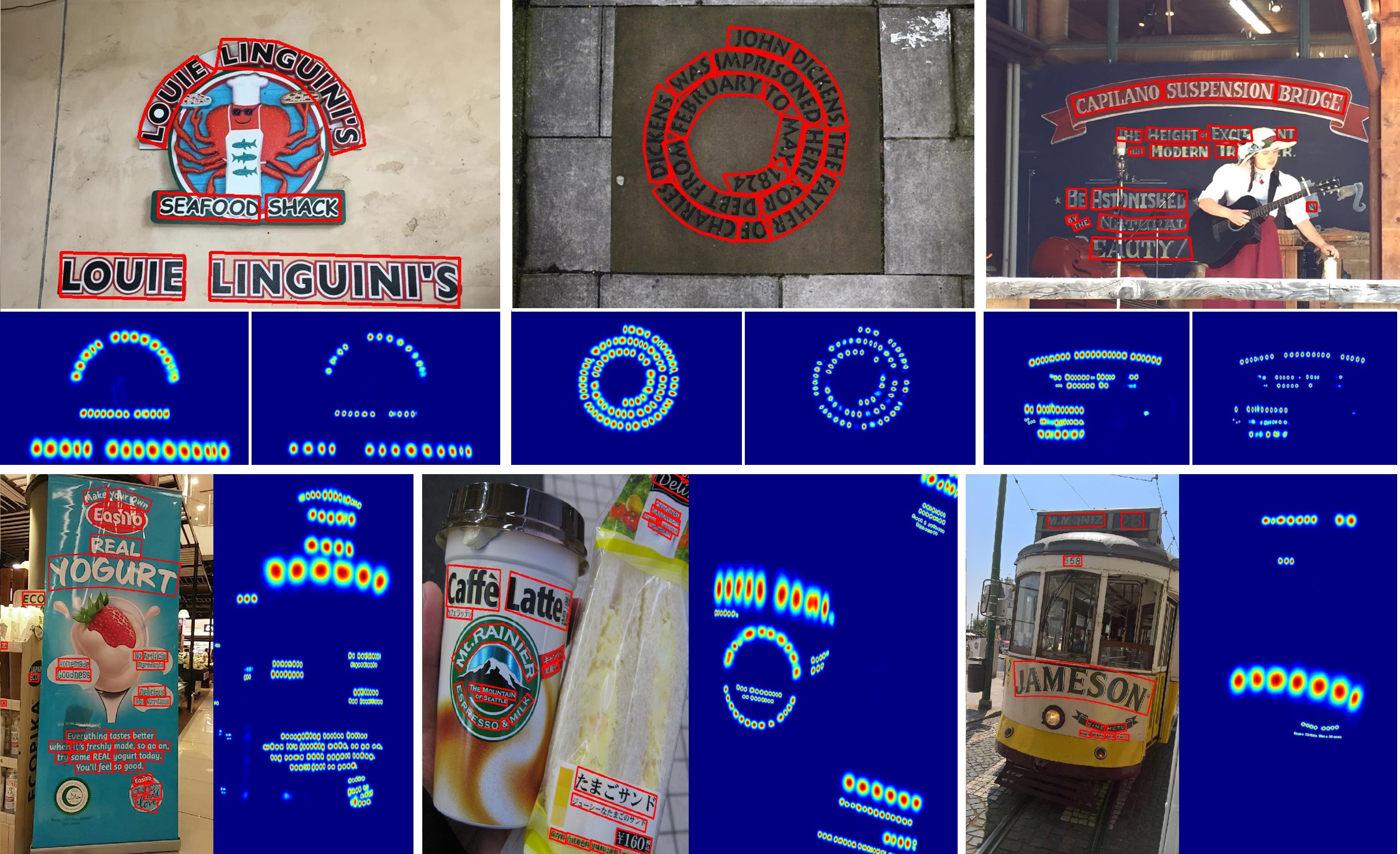}
  \caption{Results on the TotalText dataset. First row: each column shows the input image (top) with its respective region score map (bottom left) and affinity map (bottom right). Second row: each column only shows the input image (left) and its region score map (right).}
  \label{fig:result_curved} 
\end{figure*}

\subsection{Experimental Results}

\noindent\textbf{Quadrilateral-type datasets (ICDARs, and MSRA-TD500)} 
All experiments are performed with a single image resolution. The longer side of the images in IC13, IC15, IC17, and MSRA-TD500 are resized to 960, 2240, 2560, and 1600, respectively.
Table \ref{tab:result_icdar} lists the h-mean scores of various methods on ICDAR and MSRA-TD500 datasets.
To have a fair comparison with end-to-end methods, we include their detection-only results by referring to the original papers.
We achieve state-of-the-art performances on all the datasets. In addition, CRAFT runs at 8.6 FPS on IC13 dataset, which is comparatively fast, thanks to the simple yet effective post-processing.



For MSRA-TD500, annotations are provided at the line-level, including the spaces between words in the box. Therefore, a post-processing step for combining word boxes is applied. If the right side of one box and the left side of another box are close enough, the two boxes are combined together. Even though fine-tuning is not performed on the TD500 training set, CRAFT outperforms all other methods as shown in Table~\ref{tab:result_icdar}. 

\noindent\textbf{Polygon-type datasets (TotalText, CTW-1500)} 
It is challenging to directly train the model on TotalText and CTW-1500 because their annotations are in polygonal in shape, which complicates text area cropping for splitting character boxes during weakly-supervised training.
Consequently, we only used the training images from IC13 and IC17, and fine-tuning was not conducted to learn the training images provided by these datasets. At the inference step, we used the polygon generation post-processing from the \textit{region score} to cope with the provided polygon-type annotations.

The experiments for these datasets are performed with a single image resolution, too. The longer sides of the images within TotalText and CTW-1500 are resized to 1280 and 1024, respectively.
The experimental results for polygon-type datasets are shown in Table~\ref{tab:result_poly}. The individual-character localization ability of CRAFT enables us to achieve more robust and superior performance in detecting arbitrarily shaped texts compared to other methods. Particularly, the TotalText dataset has a variety of deformations, including curved texts as shown in Fig. \ref{fig:result_curved}, for which adequate inference by quadrilateral-based text detectors is infeasible. Therefore, a very limited number of methods can be evaluated on those datasets.


In the CTW-1500 dataset's case, two difficult characteristics coexist, namely annotations that are provided at the line-level and are of arbitrary polygons.
To aid CRAFT in such cases, a small link refinement network, which we call the \textit{LinkRefiner}, is used in conjunction with CRAFT. The input of the \textit{LinkRefiner} is a concatenation of the \textit{region score}, the \textit{affinity score}, and the intermediate feature map of CRAFT, and the output is a \textit{refined affinity score} adjusted for long texts. To combine characters, the \textit{refined affinity score} is used instead of the original \textit{affinity score}, then the polygon generation is performed in the same way as it was performed for TotalText. Only \textit{LinkRefiner} is trained on the CTW-1500 dataset while freezing CRAFT. The detailed implementation of \textit{LinkRefiner} is addressed in the supplementary materials.
As shown in Table~\ref{tab:result_poly}, the proposed method achieves state-of-the-art performance.

\begin{table}[t!]
  \centering
  \tabcolsep=0.15cm
  \renewcommand*{\arraystretch}{1.1}
  \begin{tabular}{c||c|c|c}
    \hline 
    \rule{0pt}{10pt} {\textbf{Method}} & {\textbf{IC13}} & {\textbf{IC15}} & {\textbf{IC17}}\\
    \hline
    \hline
    \rule{0pt}{10pt}
    Mask TextSpotter~\cite{lyu2018mask} & 91.7 & 86.0 & - \\ 
    EAA~\cite{he2018end} & 90 & 87 & - \\ 
    FOTS~\cite{liu2018fots} & 92.8 & \textcolor{black}{\textbf{89.8}} &  70.8 \\ 
    \hline
    \hline
    \rule{0pt}{10pt} \textbf{CRAFT(ours)} & \textcolor{black}{\textbf{95.2}} & {86.9} & \textcolor{black}{\textbf{73.9}} \\
    \hline
  \end{tabular}
  \vspace{3mm}
  \caption{H-mean comparison with end-to-end methods. Our method is not trained in an end-to-end manner, yet shows comparable results, or even outperforms popular methods.} 
  \label{tab:result_endtoend}
\end{table}

\subsection{Discussions}

\noindent\textbf{Robustness to Scale Variance} We solely performed single-scale experiments on all the datasets, even though the size of texts are highly diverse. This is different from the majority of other methods, which rely on multi-scale tests to handle the scale variance problem. This advantage comes from the property of our method localizing individual characters, not the whole text. The relatively small receptive field is sufficient to cover a single character in a large image, which makes CRAFT robust in detecting scale variant texts.

\noindent\textbf{Multi-language issue} The IC17 dataset contains Bangla and Arabic characters, which are not included in the synthetic text dataset. Moreover, both languages are difficult to segment into characters individually because every character is written cursively. 
Therefore, our model could not distinguish Bangla and Arabic characters as well as it does Latin, Korean, Chinese, and Japanese.
In East Asian characters' cases, they can be easily separated with a constant width, which helps train the model to high performance via weakly-supervision.



\noindent\textbf{\bf Comparison with End-to-end methods} 
Our method is trained with the ground truth boxes only for detection, but it is comparable with other end-to-end methods, as shown in Table.~\ref{tab:result_endtoend}. From the analysis of failure cases, 
we expect our model to benefit from the recognition results, especially when the ground truth words are separated by semantics, rather than visual cues.



\noindent\textbf{Generalization ability}
Our method achieved state-of-the-art performances on 3 different datasets without additional fine-tuning. This demonstrates that our model is capable of capturing general characteristics of texts, rather than overfitting to a particular dataset.


\section{Conclusion}
We have proposed a novel text detector called CRAFT, which can detect individual characters even when character-level annotations are not given. The proposed method provides the \textit{character region score} and the \textit{character affinity score} that, together, fully cover various text shapes in a bottom-up manner. Since real datasets provided with character-level annotations are rare, we proposed a weakly-supervised learning method that generates pseudo-ground truthes from an interim model.
CRAFT shows state-of-the-art performances on most public datasets and demonstrates generalization ability by showing these performances without fine-tuning. 
As our future work, we hope to train our model with a recognition model in an end-to-end fashion to see whether the performance, robustness, and generalizability of CRAFT translates to a better scene text spotting system that can be applied in more general settings.


\vspace{2mm}
\noindent\textbf{Acknowledgements.} The authors would like to thank Beomyoung Kim, Daehyun Nam, and Donghyun Kim for helping with extensive experiments.

{\small
\bibliographystyle{ieee}
\bibliography{mybib}
}

\clearpage
\appendix

\section{\textit{LinkRefiner} for CTW-1500 dataset}

\begin{figure}[h]
	\begin{center}
        \includegraphics*[width=0.8\linewidth, clip=true]{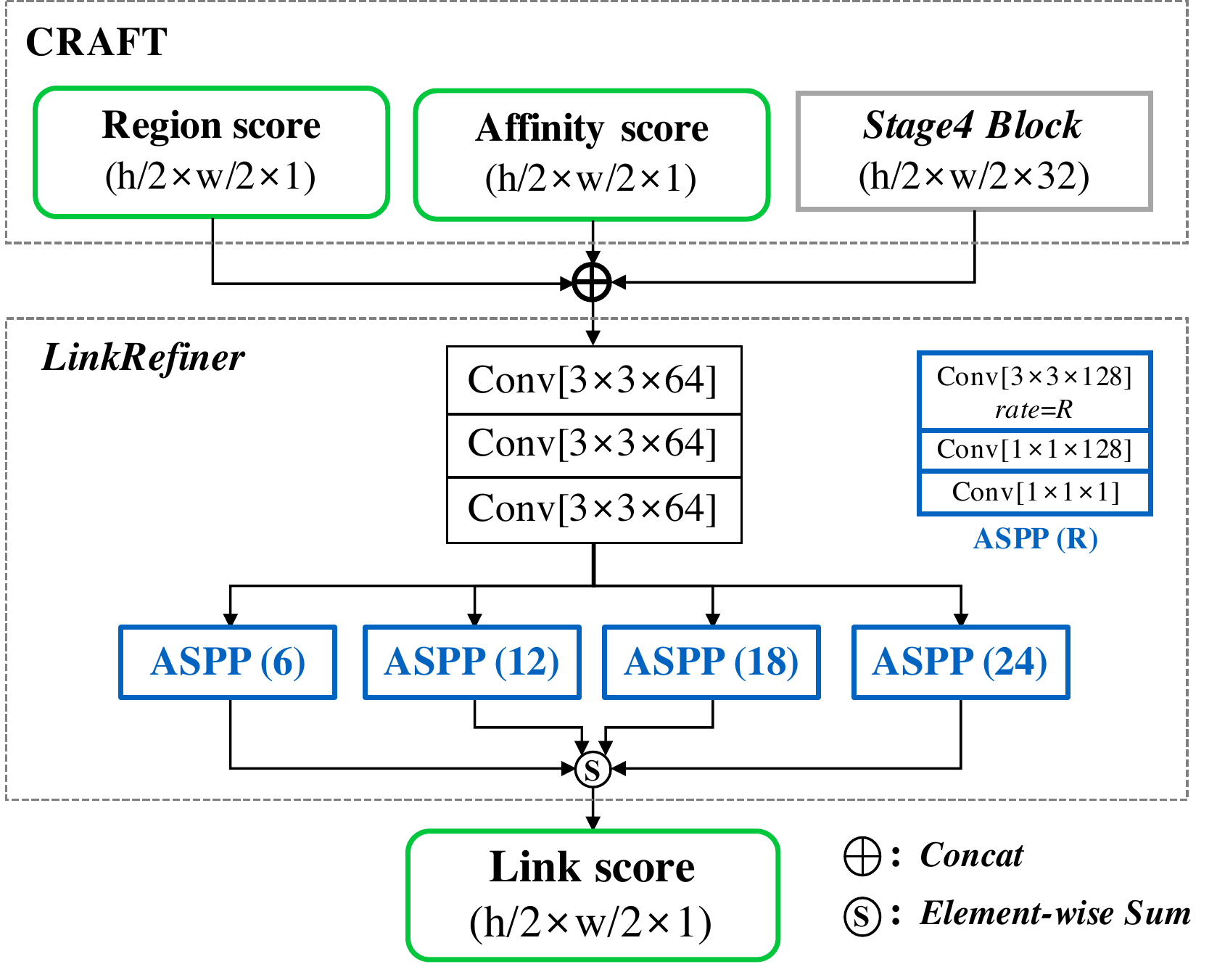}
         \vspace{+1mm}
        \caption{Schematic illustration of \textit{LinkRefiner} architecture.}
  	    \label{fig:architecture_refiner} 
    \end{center}
     \vspace{-1.5mm}
\end{figure}

CTW-1500 dataset~\cite{yuliang2017detecting} provides polygon-only annotations without text transcriptions. 
Furthermore, annotations of CTW-1500 are provided at the line-level and does not consider spaces as separation cues. This is far from our assumption of affinity, which is that the score for affinity is zero for characters with a space between them.

To obtain a single-long polygon from the detected characters, we employ a shallow network for link refinement,  so called \textit{LinkRefiner}. The architecture of the \textit{LinkRefiner} is shown in Fig.~\ref{fig:architecture_refiner}. The input of the \textit{LinkRefiner} is a concatenation of the \textit{region score}, the \textit{affinity score}, and the intermediate feature map from the network, that is the output of \textit{Stage4} of the original CRAFT model. Atrous Spatial Pyramid Pooling (ASPP) in \cite{chen2018deeplab} is adopted to ensure a large receptive field for combining distant characters and words onto the same text line.

\begin{figure}[t]
	\begin{center}
        \includegraphics*[width=0.8\linewidth, clip=true]{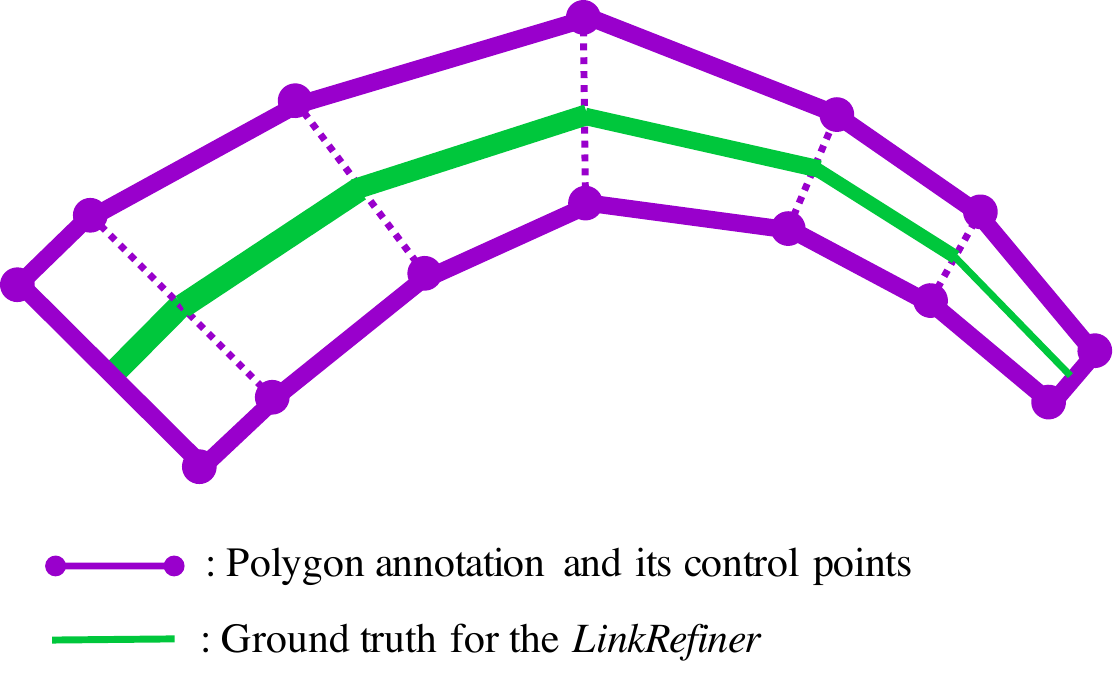}
        \caption{Ground truth generation for \textit{LinkRefiner}.}
  	    \label{fig:gt_generation_refiner} 
    \end{center}
    \vspace{-1.5mm}
\end{figure}

For the ground truth of the \textit{LinkRefiner}, lines are simply drawn between the centers of the paired control points of the annotated polygons, which is similar to the text line generation used in \cite{he2016accurate}. The width of each line is proportional to the distance between paired control points. The ground truth generation for the \textit{LinkRefiner} is illustrated in Fig.~\ref{fig:gt_generation_refiner}. The output of the model is called the \textit{link score}. For training, only the \textit{LinkRefiner} is trained on the CTW-1500 training dataset, while freezing CRAFT.

After training, we have the outputs produced by the model, which are the \textit{region score}, the \textit{affinity score}, and the \textit{link score}. Here, the  \textit{link score} is used instead of the original \textit{affinity score}, and the text polygon is obtained entirely through the same process as done with TotalText. The CRAFT model localizes the individual characters, and the \textit{LinkRefiner} model combines the characters as well as the words separated by spaces, which are required by the CTW-1500 evaluation.

\begin{figure*}[t]
  \centering
  \includegraphics*[width=16.5cm, clip=true]{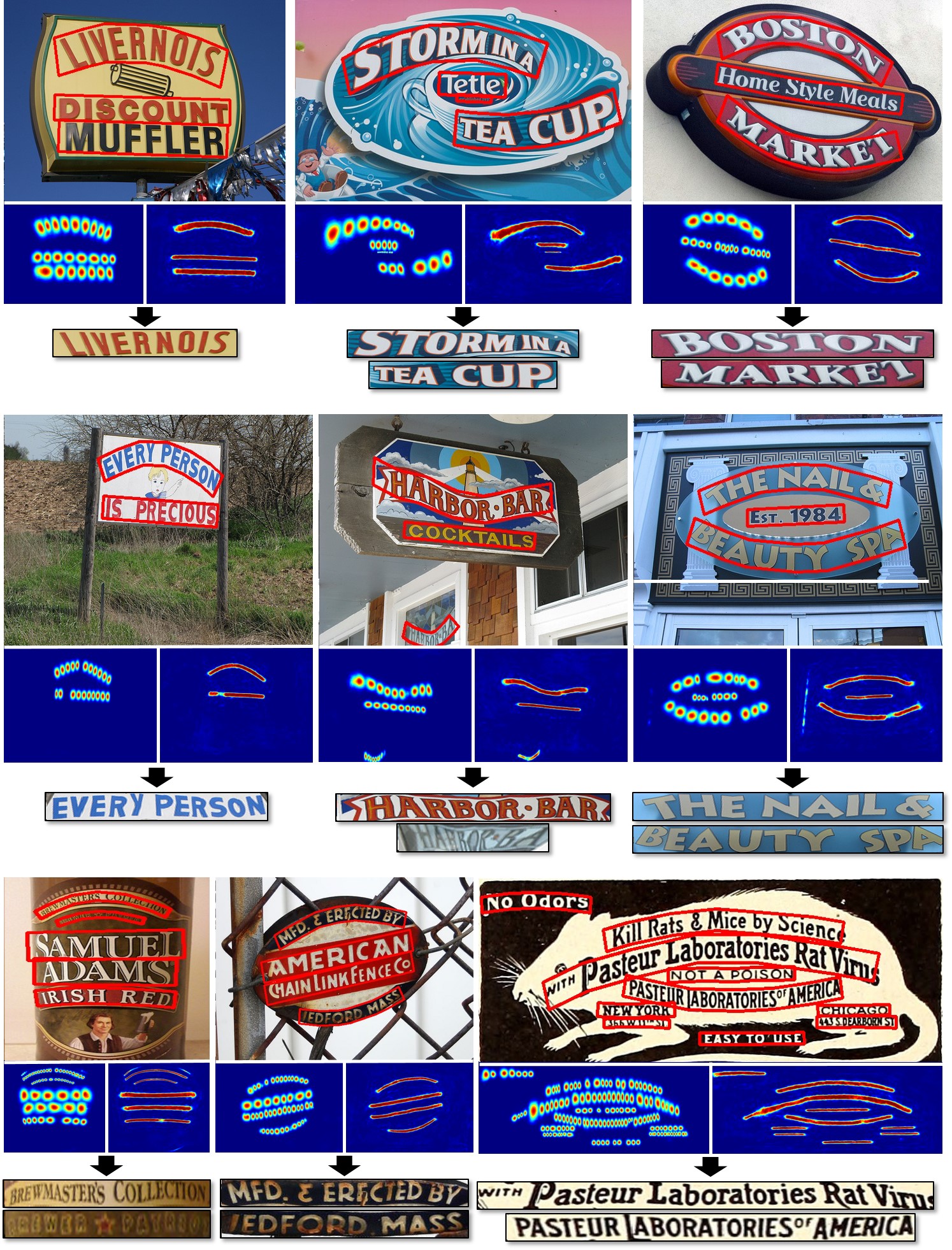}
  \caption{Results on CTW-1500 dataset. For each cluster: the input image (top), region score (middle left), link score (middle right), and the resulting rectified polygons for curved texts (bottom, below arrow) are shown. Note that the affinity scores are not rendered and are unused in the CTW-1500 dataset.} 
  \label{fig:result_ctw} 
\end{figure*}

The results on the CTW-1500 dataset are shown in Fig.~\ref{fig:result_ctw}. Very challenging image samples with long and curved texts are successfully detected by the proposed method. Moreover, with our polygon representation, the curved images can be rectified into straight text images, which are also shown in Fig.~\ref{fig:result_ctw}. We believe this ability for rectification can further be of use for recognition tasks.

\end{document}